\documentclass{article}

\usepackage{arxiv}
\usepackage{tikz}
\usetikzlibrary{shapes.geometric, arrows.meta, positioning}

\tikzstyle{block} = [rectangle, rounded corners, minimum width=3.2cm, minimum height=1.2cm, text centered, draw=black, fill=gray!10]
\tikzstyle{arrow} = [thick, -{Latex[width=2mm,length=2mm]}]

\usepackage[utf8]{inputenc}     
\usepackage[T1]{fontenc}        
\usepackage{hyperref}           
\usepackage{url}                
\usepackage{booktabs}           
\usepackage{amsfonts}           
\usepackage{nicefrac}           
\usepackage{microtype}          
\usepackage{siunitx}            
\usepackage{graphicx}           
\usepackage{amsmath}            
\usepackage{csquotes}
\usepackage[style=numeric,backend=biber,sorting=none]{biblatex}
\usepackage{doi}                
\usepackage{lipsum}
\usepackage{enumitem}
\usepackage{caption} 
\usepackage{pgfplots}
\pgfplotsset{compat=1.18}
\usepackage{pdflscape}
\usepackage{afterpage}
\usepackage{capt-of} 
\usepackage{fancyhdr}
\usepackage[export]{adjustbox}
\usepackage[all]{hypcap}
\usepackage{lscape}
\usepackage{geometry}
\usepackage{rotating}
\usepackage[justification=centering, singlelinecheck=false]{caption}

\setlist[itemize]{noitemsep, topsep=0pt, parsep=0pt, partopsep=0pt}
\title{Towards Resource-Efficient Multimodal Intelligence: Learned Routing among Specialized Expert Models}

\author{%
  \href{https://orcid.org/0009-0002-9443-5621}{\includegraphics[scale=0.06]{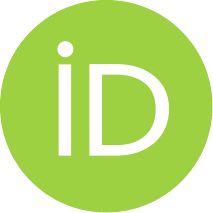}\hspace{1mm}Mayank Saini} \\
  PwC US \\
  \texttt{mayank.s.saini@pwc.com} \\
  \And
  \href{https://orcid.org/0000-0001-8064-2035}{\includegraphics[scale=0.06]{orcid.pdf}\hspace{1mm}Arit Kumar Bishwas}\thanks{$\ast$Corresponding author} \\
  PwC US \\
  \texttt{arit.kumar.bishwas@pwc.com}
}

\hypersetup{
  pdftitle={Towards Resource-Efficient Multimodal Intelligence: Learned Routing among Specialized Expert Models},
  pdfsubject={AI, Multimodal Reasoning, LLM Routing},
  pdfauthor={Mayank Saini, Arit Kumar Bishwas},
  pdfkeywords={LLM Routing, Multimodal AI, Optimal Model Selection, AI Orchestration, Resource-Efficient Deployment}
}

\usepackage{float}
\begin{document}
\maketitle
\begin{abstract}
As AI moves beyond text, large language models (LLMs) increasingly power vision, audio, and document understanding; however, their high inference costs hinder real-time, scalable deployment. Conversely, smaller open-source models offer cost advantages but struggle with complex or multimodal queries. We introduce a unified, modular framework that intelligently routes each query - textual, multimodal, or complex - to the most fitting expert model, using a learned routing network that balances cost and quality. For vision tasks, we employ a two-stage open-source pipeline optimized for efficiency and reviving efficient classical vision components where they remain SOTA for sub-tasks. On benchmarks such as Massive Multitask Language Understanding (MMLU) and Visual Question Answering (VQA), we match or exceed the performance of always-premium LLM (monolithic systems with one model serving all query types) performance, yet reduce the reliance on costly models by over 67\%. With its extensible, multi-agent orchestration, we deliver high-quality, resource-efficient AI at scale.
\end{abstract}


\section{Introduction}

\begin{figure}[htbp]
  \centering
  \includegraphics[width=0.95\linewidth]{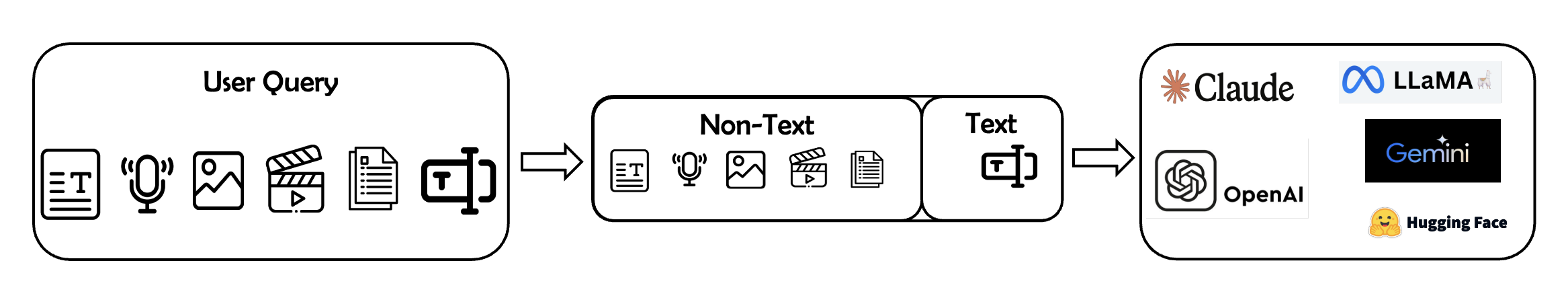}
  \caption{\small\textit{Initial modality-based routing where text queries are routed to LLMs based on complexity, while non-text inputs (e.g., image, audio, video, document) are dispatched to specialized expert pipelines.}}
  \label{fig:routing}
\end{figure}

Large language models (LLMs) have demonstrated strong performance on a wide spectrum of tasks including open-domain dialogue \cite{roller2021recipes}, code generation \cite{chen2021evaluating}, and document understanding \cite{lin2024zendb}. Their generative fluency and zero-shot capabilities make them particularly appealing for end-to-end AI assistants \cite{radford2019language}. However, the high computational cost and latency associated with these models pose a significant barrier to large-scale, real-time deployment—especially when used uniformly across all incoming queries \cite{zhou2024efficientinf}. In practice, most tasks do not require the full capacity of the largest models, and indiscriminate usage leads to unnecessary resource expenditure \cite{treviso2023efficient}.
This paper proposes that routing queries across a pool of specialized, cost-calibrated experts—based on task type, modality, and predicted complexity—offers a scalable and efficient alternative to monolithic LLM deployment. We formalize our architecture (Figure~\ref{fig:routing}), which begins by decomposing a user query into its underlying modality—text, image, audio, video, or document—followed by intelligent branching into textual or non-text pipelines. Each branch routes the query to the most suitable model from a configurable expert pool, including Claude, LLaMA, Gemini, OpenAI, and various open-source models. This structured routing paradigm enables our approach to generalize across tasks without sacrificing operational efficiency.

In the current landscape, AI assistants must increasingly handle multimodal queries—not just text, but also supporting media inputs such as screenshots, voice clips, scanned documents, and short videos. This shift introduces new orchestration challenges: models that excel in one modality may not generalize well to others, and large multimodal LLMs are prohibitively expensive when used universally \cite{expert_orchestration_2025}. Some frameworks rely on centralized orchestrators, like HuggingGPT \cite{shen2023hugginggpt}, which invoke specialized tools under the supervision of a large LLM. Others, like FrugalGPT \cite{chen2023frugalgpt}, employ adaptive routing strategies to direct queries to domain-specific agents based on cost-performance trade-offs. However, many such systems either rely too heavily on the central controller or fail to dynamically optimize routing based on cost-performance considerations.

We introduce a modular execution graph that begins with modality classification, followed by cost-sensitive model selection guided by a tunable threshold-based routing function. Queries predicted to be simple are handled by efficient open-source models; more complex or ambiguous ones are escalated to high-performance LLMs. For visual reasoning, the framework incorporates Couplet, a modular vision-language pipeline that combines vision encoders (e.g., CLIP \cite{radford2021learning}, ResNet \cite{he2016deep}) with compact generative models (e.g., Phi-3 \cite{phi3_technical2024}, Mistral-7B \cite{mistral7b_2023}). This modular fusion enables high-quality image analysis by leveraging traditional models which remain state-of-the-art for specific visual tasks. By doing so, we avoid invoking computationally intensive, end-to-end vision LLMs unless explicitly required. Additionally, we integrate robust fallback mechanisms and a multi-agent orchestration layer to efficiently manage complex, cross-modal scenarios. This design makes the framework highly suitable for scalable deployment in resource-constrained environments.

\section{Related Work}
\label{sec:related}
Recent advances in large language models (LLMs) and multimodal intelligence enable reasoning across text, images, audio, video, and documents, yet indiscriminate use of the strongest models imposes substantial cost. This has driven research on cost-aware routing among model pools, orchestration mechanisms that coordinate specialized components, and agent frameworks for decomposing complex requests under practical latency and budget constraints.

Selective routing among LLMs aims to preserve accuracy while reducing premium calls. \emph{HybridLLM} \cite{ding2024hybridllm} verifies the output of a smaller model and escalates uncertain cases to a stronger model, reducing large-model usage by roughly 40\% at the expense of additional cascade latency. \emph{RouteLLM} \cite{ong2024routellm} trains a classifier on human preference data to predict when a query benefits from a large model, yielding over 50\% savings in expensive calls while maintaining strong accuracy on GSM8K \cite{gsm8k_cobbe2021} and MT-Bench \cite{mtbench_acl2024}. \emph{FrugalGPT} \cite{chen2023frugalgpt} learns query-dependent cascades under budget constraints, combining prompt adaptation and model approximation to cut cost by as much as 98\%. However, no existing architecture addresses the complete multimodal routing challenge. RouteLLM and HybridLLM handle only text routing, while HuggingGPT provides multimodal orchestration but lacks efficient routing mechanisms. We introduce the first unified architecture that combines intelligent routing across all modalities with selective orchestration—a capability gap that current systems cannot address.

Moving beyond text-only settings, multimodal orchestration investigates how to dispatch and integrate modality-specific tools. \emph{HuggingGPT} \cite{shen2023hugginggpt} uses a powerful controller LLM to decompose requests and invoke domain experts (e.g., captioning, ASR), integrating their outputs into a final response; although broadly capable, it keeps a heavyweight controller in the loop even for simple subtasks. Instead, we apply embedding-based semantic matching (Route0x \cite{damodaran2024route0x}) to map non-text queries directly to vision, audio, or document pipelines, avoiding controller calls when not necessary. For vision, a “Couplet” path pairs traditional detectors (e.g., CLIP/YOLO-style components) with a small-model (SLM) reasoning layer to balance cost and accuracy without invoking a large controller by default.

General agent frameworks provide reusable mechanisms for task decomposition and coordination. Toolkits such as LangChain \cite{chase2023langchain}, AutoGen \cite{li2023autogen}, SmolAgents \cite{hf_smolagents}, LangGraph \cite{hf_langgraph}, and Atomic Agents \cite{atomicagents} support breaking requests into subtasks and composing outputs, enabling sequences like optical character recognition (OCR) $\rightarrow$ translation $\rightarrow$ summarization for document workflows. While effective on multi-step problems, these frameworks can introduce orchestration overhead on simple cases. We activate multi-agent orchestration selectively for truly complex or multi-dependency queries and aggregate sub-results via a mixture-of-experts coordinator, consistent with evidence that compact SLMs can be competitive on many agentic subtasks \cite{belcak2025slm}.

We unify these perspectives by combining text-specific complexity analysis with universal semantic similarity routing—embedding-based dispatch to the closest task exemplar—and modality-aware pipelines—specialized pipelines per modality (text, vision, audio, document)—invoking multi-agent coordination only when the problem structure requires it. This positioning preserves the efficiency benefits of selective routing while supporting practical multimodal deployments.

\section{Framework}
This section presents our framework for cost-aware, modality-specific execution across text, image, audio, video, and document inputs. We outline the architectural principles and routing mechanisms that govern intake, classification, and model selection under user budget and policy constraints. The subsequent subsections describe the high-level system layout and component workflow, setting up the empirical evaluation that follows.

\subsection{High-Level System Architecture}
We present a unified, modular, and cost-aware AI architecture designed for dynamic orchestration across diverse input types and task complexities. The system seamlessly processes text, image, audio, video, and document inputs through a streamlined user interface, and intelligently routes them to optimized model pipelines.
\begin{figure}[htbp]
  \centering
  \includegraphics[width=\textwidth, keepaspectratio]{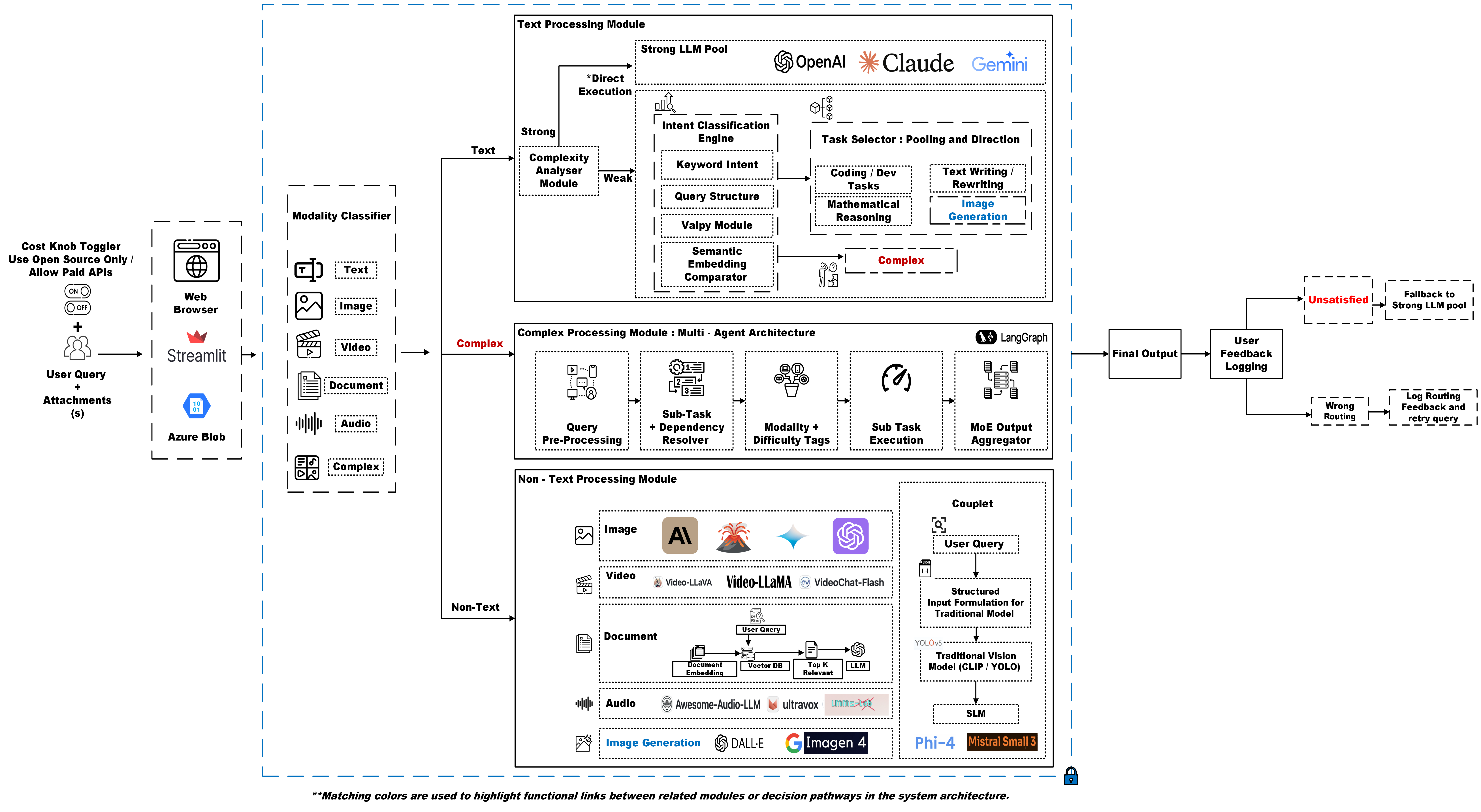}
  \caption{\small\textit{High-Level System Architecture.}}
  \label{fig:high_level_arch}
\end{figure}
As illustrated in Figure~\ref{fig:high_level_arch}, the pipeline begins with the {Streamlit-based user interface}, which accepts natural language queries along with optional attachments (e.g., images, audio, video and documents). The {Modality Classifier} rapidly categorizes the input as text, non-text, or complex (multimodal), triggering appropriate downstream modules.

{User preference constraints}—such as restricting access to closed-source, paid APIs (e.g., GPT-4, Gemini, DALL·E)—are specified through a toggle interface and propagated as global routing parameters across the execution graph. These constraints affect two key decision points—LLM selection and modality-specific execution. At the LLM selection stage, user queries are routed through model pools that respect the configured budget preference, and premium models are invoked only when explicitly permitted. For modality-specific execution on image, audio, and document tasks, it dynamically chooses between high-performing open-source models (e.g., Qwen-VL, DeepSeek, CLIP) and paid APIs (e.g. GPT-4 Vision, Gemini Pro) according to the toggle state.

Crucially, these constraints are applied without compromising response quality. Our task-aware router substitutes general-purpose LLMs with domain-optimized open models—such as Qwen-Code for programming or Mistral for summarization—ensuring comparable output fidelity at significantly lower cost. This enables deployment flexibility across both enterprise-scale and resource-constrained settings.

\noindent With preference and modality policies in place, we proceed to dynamically route queries through specialized branches. {Text queries} are passed to the Complexity Analyzer Module, which determines whether to route them directly to a {Strong LLM Pool} (e.g., GPT-4, Claude, Gemini) or through an {Intent Classification Engine}. This engine combines keyword detection, structural analysis, semantic embedding, and the Valpy prompt-classifier~\cite{valpy} to robustly detect and direct weak queries to task-specific pipelines (such as coding, summarization, or mathematical reasoning).
If the system cannot resolve the intent or multiple subtasks are detected, the query is flagged as {Complex} and escalated to the {LangGraph-based Multi-Agent System}.
{Complex queries} undergo query decomposition and dependency resolution, where each subtask is tagged with its modality and difficulty level. These are routed to specialized agents via the {Agent Dispatcher}, and results are compiled using a {Mixture-of-Experts (MoE) Aggregator} to ensure accurate, unified output.

Non-text queries are routed directly to specialized domain-specific modules. The Image Module handles vision tasks and supports various models, including open-source solutions such as Qwen-VL vision models and premium alternatives like GPT-4. The Video Module integrates vision-language assistants tailored for video analysis, such as Video-LLaVA~\cite{wang2023videollava}, VideoChat-Flash~\cite{li2025videochatflash}, and Video-LLaMA2, a spatio-temporal video language model~\cite{videollama2_2024}. The Document Module employs document embedding approaches combined with retrieval-augmented generation (RAG) pipelines. For audio inputs, the Audio Module leverages dedicated speech-to-text (STT) and LLM-hybrid models, including Ultravox~\cite{ultravox} and Awesome-Audio-LLM~\cite{awesome_audio_llm}. The Image Generation Module manages text-to-image generation tasks using models such as DALL·E and Imagen. Lastly, the Couplet Subsystem applies structured prompting to classical machine learning models (e.g., YOLO~\cite{redmon2016you}, DETR~\cite{detr}), bridging them with Small Language Models (SLMs) to produce coherent user responses.

Follow-up text queries triggered after non‑text module interactions are handled by a separately instantiated Mixture‑of‑Experts (MoE) model—in our case, Mixtral 8×7B from Mistral AI~\cite{jiang2024mixtral}. This dedicated MoE is loaded with instructions to preserve context from the prior multimodal query, ensuring that follow-ups are contextually grounded without redundant re‑analysis of the original input.

{User feedback} is collected post-inference and categorized based on the type of issue encountered. If a response is marked as \textit{unsatisfactory}, the system triggers a fallback to a stronger LLM (e.g., GPT-4, Claude) for immediate reprocessing. If the feedback indicates a \textit{routing error}—such as incorrect model or modality, we log the misrouting event and reattempt the query using an updated route.

\subsection{Detailed Workflow and Components}

This subsection details the end-to-end workflow and component interfaces that transform user inputs into executable routes. We formalize modality identification, text-only complexity estimation, semantic similarity mapping, and the routing objective that selects efficient or premium paths under policy constraints. We then outline multi-agent decomposition, memory management, and feedback-driven policy adjustment that together ensure scalable and reliable operation of our framework.

\subsubsection{Query Intake, Classification, and Modality Flagging}
We perform intake preprocessing for modality identification using attachment detection, MIME/extension checks, audio recognition (with transcription when needed), and semantic embedding plus keyword classifiers. Each incoming request is assigned to one of four input types (independent text, text with attachments, independent attachments, or audio recordings), then classified by modality (text, audio, video, image, document, or multimodal). For downstream routing, it is mapped to one of four execution categories—text-only, non-text, hybrid, or follow-up—under the user’s cost preferences (paid vs open-source, or automatic adaptation).

Modality classification follows a hierarchical decision tree. Primary classification uses MIME type analysis with 15 predefined categories (image/*, audio/*, video/*, application/pdf, text/*). Secondary classification employs file extension mapping with fallback rules for ambiguous cases. Tertiary classification applies content-based analysis using magic number detection for binary files and encoding validation for text files. Audio inputs trigger automatic speech-to-text transcription using audio transcription models before text pipeline routing. The attachment detector integrates MIME validation, file signature verification, and content-length analysis to ensure robust modality identification across diverse input formats.

Once classified, routing aligns with detected modality and structure. For text-only queries, a lightweight complexity classifier estimates whether to remain on efficient models or escalate. We compute three features—\(\mathcal{I}(Q)\) (intent/task alignment), \(\mathcal{L}(Q)\) (length/verbosity) and \(\mathcal{S}(Q)\) (semantic/structural density) and obtain a class distribution \(P(c \mid f(Q))\) over \(c \in \{\text{low},\text{medium},\text{high}\}\) via a RouteLLM-style predictor.
In parallel, a scalar difficulty score is formed as
\begin{equation}
C(Q) = \alpha \cdot \mathcal{I}(Q) + \beta \cdot \mathcal{L}(Q) + \gamma \cdot \mathcal{S}(Q),
\end{equation}
where $\mathcal{I}(Q)$ represents alignment with task archetypes such as summarization or coding, $\mathcal{L}(Q)$ denotes linguistic length or verbosity, and $\mathcal{S}(Q)$ captures semantic and structural complexity. 

The complexity features are computed algorithmically as follows. Intent alignment $\mathcal{I}(Q)$ is calculated using weighted keyword matching, where each task category (coding, mathematics, summarization) has predefined keyword dictionaries with empirically determined weights ranging from 1–5 based on semantic importance. Linguistic complexity $\mathcal{L}(Q)$ incorporates text length normalization, syntactic depth analysis, and vocabulary diversity metrics using Shannon entropy. Structural complexity $\mathcal{S}(Q)$ employs a multi-stage analysis: first, Abstract Syntax Tree (AST) parsing for code detection with 94\% accuracy on programming queries; second, SQL statement validation using parse tree analysis; third, lexical analysis using context-free grammar recognition for 25+ programming languages. Fourth, regex pattern matching captures mathematical notation, structured data, and domain-specific formats. The threshold $\tau$ is dynamically calibrated on a held-out validation set by sweeping candidate values and selecting the operating point that maximizes routing accuracy under a fixed cost budget; the boundary is re-estimated periodically as data distribution and deployment constraints evolve.

The complexity score $C(Q)$ is calibrated to maintain monotonicity with $P(c=\text{high}\mid f(Q))$, with $\alpha, \beta,$ and $\gamma$ treated as tunable parameters. Queries satisfying the condition $C(Q) < \tau$ are routed to efficient open-source model pools (e.g., QwenCoder, WizardMath). Queries exceeding this threshold escalate to premium LLMs (e.g., GPT-4, Claude, Gemini). In cases involving ambiguity or multi-step dependencies, queries are deferred to a planner.

Non-text inputs (images, audio, video, documents) and hybrid requests (text plus attachments or multiple attachment types) bypass complexity scoring and are dispatched directly to modality-specific pipelines, since perceptual decoding and multimodal embedding impose a substantial baseline cost. Selection among open-source or closed-source APIs follows user preferences. Vision requests may take a Couplet path that fuses traditional perception modules with a small-model reasoning layer; documents invoke OCR/structure extraction; audio uses ASR; and video employs key-frame or track extraction prior to reasoning. When semantic analysis indicates interdependent subtasks, a LangGraph-based multi-agent planner decomposes the request and executes components in parallel or sequence as appropriate, and a mixture-of-experts (MoE) aggregator merges sub-results.

We make complexity explicit for text because it governs predictable escalation thresholds and avoids overspending on simple instructions.
For non‑text, the baseline perceptual cost is already high, so modality\-aware dispatch and semantic similarity are the primary routing signals rather than an efficient\-versus\-complex branch.

\subsubsection{Routing and Orchestration}
\label{sec:routing-orchestration}

We use a centralized orchestration engine that assigns each query—according to its downstream processing type (text-only, non-text, hybrid, or follow-up)—to an execution route based on modality alignment, task affinity, cost profile, and user-defined preferences. This routing layer balances the structural characteristics of the query with system constraints and user directives, ensuring that each request is handled in a cost-efficient and semantically aligned manner.
\begin{equation}
\label{eq:routing-objective}
R(Q) = \arg\max_{r \in \mathcal{R}} \left[ \delta_m \cdot S_m(Q, r) + \delta_u \cdot S_u(Q, r) + \delta_t \cdot S_t(Q, r) - \lambda_c \cdot C_r(r) \right],
\end{equation}
In Eq.~\ref{eq:routing-objective}, $\mathcal{R}$ denotes the space of all available execution routes, including efficient model paths, premium model paths, or multi-agent pipelines. The modality alignment score, $S_m(Q, r)$, measures how well a given route accommodates the input modality or modalities of the query. The user preference match score, $S_u(Q, r)$, evaluates the degree to which a selected route aligns with user-specified constraints, such as preferences for open-source or premium models. The task-type affinity score, $S_t(Q, r)$, quantifies the suitability of the route for handling the query's task type. Finally, $C_r(r)$ represents the inherent resource cost associated with executing the chosen route. The parameters $\delta_m$, $\delta_u$, $\delta_t$, and $\lambda_c$ are tunable routing weights controlling the influence of each factor on route selection.

These routing weights are empirically derived through a hierarchical calibration process. Modality alignment receives the highest priority because a modality mismatch can lead to complete task failure, establishing $\delta_m$ as the dominant coefficient. User‑preference alignment is weighted next to guarantee deployment feasibility and cost compliance across organizational policies. Task‑type affinity follows, since many model–task mismatches can be mitigated through adaptive prompting. Finally, the cost penalty coefficient $\lambda_c$ is determined via Pareto‑front analysis that balances accuracy against resource expenditure, yielding an optimal performance–cost trade‑off. All four weights are then $\ell_1$‑normalized to ensure consistent scaling while preserving this priority hierarchy across deployment scenarios.

The following sections describe how this orchestration logic is concretely applied across both single-modality (text) and complex multimodal inputs.

\subsubsection{Text Query Routing}
Text-based queries are routed through a multi-stage classification stack to determine both complexity and intent. The system distinguishes between lightweight (e.g., summarization, rephrasing) and heavyweight (e.g., mathematical reasoning, multi-step coding) tasks using a hybrid of syntactic, semantic, and structural signals.

The intent classification engine comprises four components. We begin with a keyword-based filter that scans for high-confidence terms (e.g., "generate code", "solve integral", "summarize", "translate") and, when confidence exceeds a threshold, triggers direct routing to a specialized model. Next, the analysis considers structure—code blocks, numerical formulas, document-style formatting, and stepwise instructions—to separate coding, computation, or NLP pipelines. Subsequently, a Valpy classifier~\cite{valpy} performs zero-shot intent classification and maps fine-grained labels (e.g., “mathematical reasoning”, “coding”, “summarization”) to task categories. Finally, a semantic similarity mapper matches query embeddings against task cluster centroids (e.g., “creative generation”, “formal reasoning”, “document QA”) using fine-tuned transformer embeddings with FAISS-based indexing.

Based on the classifier outputs, the orchestration logic assigns each query to one of three execution paths. 
Simple queries are routed to efficient, open-source models like QwenCoder for coding queries or WizardMath for analytical queries. 
More complex queries that require deeper reasoning or richer context are handled by commercial models like GPT-4, Claude, or Gemini Pro. 
When classification confidence is low or the modality is unclear, the system invokes a multi-agent planner that performs staged task decomposition followed by adaptive model selection.

In special cases like text-to-image generation, routing redirects to a non-text module with modality handoff. If the query is flagged as a follow-up (based on SLM + semantic tracer), context is passed forward and routed to a Mixture-of-Experts model with retained modality-awareness.

{Routing Equation for Text Queries:}
\begin{equation}
R_{\text{text}}(Q) = 
\begin{cases}
    M_{\text{eff}}(Q) & \text{if } C(Q) < \tau \text{ and user prefers efficient} \\
    M_{\text{prem}}(Q) & \text{if } C(Q) \geq \tau \text{ or task flagged complex} \\
    \text{Agent Cascade}(Q) & \text{otherwise (fallback to multi-agent system)}
\end{cases}
\end{equation}
Here, $C(Q)$ is the query complexity score, $\tau$ is a dynamic complexity threshold, and $M_{\text{eff}}, M_{\text{prem}}$ denote model classes aligned to efficiency or premium capability.

\subsubsection{Complex Multimodal and Multi-Agent System}
\begin{figure}[htbp]
\centering
\includegraphics[width=0.95\textwidth]{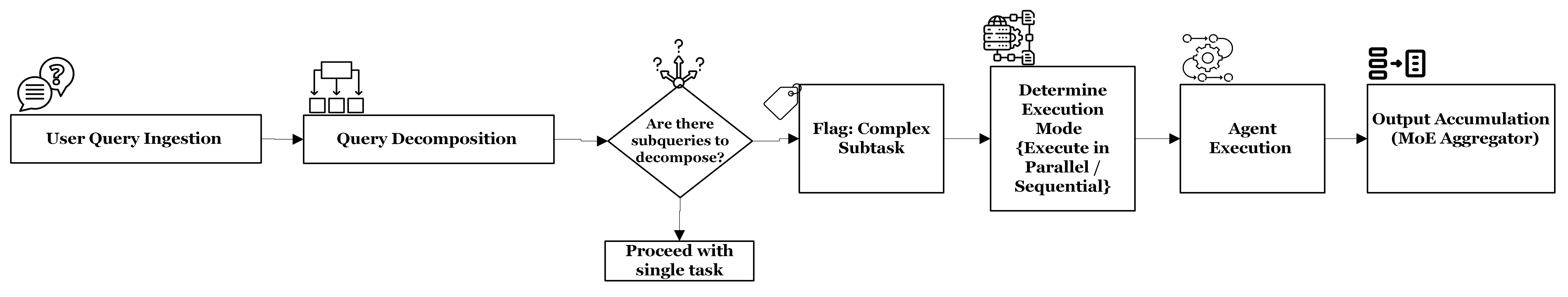}
\caption{\small\textit{LangGraph Multi-Agent System Workflow.}}
\label{fig:multi_agent}
\end{figure}

Complex queries involving multiple modalities—text, image, audio, video, or structured documents—are routed to a modular \emph{LangGraph-powered Multi-Agent System}. Each query is decomposed into atomic subtasks aligned to modality and task-type, and dispatched to appropriate agents with scoped context. Figure~\ref{fig:multi_agent} summarizes the resulting execution graph, including query decomposition, dispatch to modality-specific agents, and the flow of intermediate results.

Each LangGraph node specifies its agent type (Vision, NLP, Audio, Document, or Multimodal Fusion), enumerates its capabilities (structured document analysis, visual understanding, audio transcription and sentiment analysis, mathematical and logical reasoning, and multi-step generative workflows across modalities), and defines its context scope via a local task buffer with optional injection of compressed global state when required.

Execution logic follows either sequential pipelines when dependencies exist or parallel branches for independent subtasks. Query decomposition adheres to a modality/task DAG inferred at runtime, and downstream synthesis is handled by a Mixture-of-Experts (MoE) executor—an aggregation layer that blends component outputs using non-negative weights that sum to one— that merges partial outputs into a cohesive, cross-modal response.

Concretely, the aggregator combines the four component outputs by first mapping each node’s output $o_i$ into a shared latent space through a lightweight modality adapter $A_{m_i}$, producing $\tilde{o}_i = A_{m_i}(o_i)$. A scalar relevance score $s_i$ is then computed for each component using signals already defined in the system—modality alignment $S_m(Q,i)$, task affinity $S_t(Q,i)$, and the component’s own confidence $\mathrm{conf}_i$—via a nonnegative combination
\[
s_i \;=\; \alpha_m\,S_m(Q,i)\;+\;\alpha_t\,S_t(Q,i)\;+\;\alpha_c\,\mathrm{conf}_i,
\]

The coefficients $(\alpha_m,\alpha_t,\alpha_c)$ are tuned using the performance-driven procedure described in Section~\ref{sec:routing-orchestration} and rescaled to a common range for stability; no fixed values are assumed. These scores are converted into normalized weights
\begin{equation}
w_i \;=\; \frac{\exp(s_i)}{\sum_{j}\exp(s_j)} \qquad \text{with} \quad \sum_i w_i = 1,
\end{equation}
and the fused representation is obtained by a weighted sum
\begin{equation}
\tilde{o}_{\mathrm{agg}} \;=\; \sum_{i} w_i\,\tilde{o}_i.
\end{equation}
If one component clearly dominates (according to an empirically calibrated dominance criterion), the system selects that output directly; otherwise, the weighted fusion is decoded to text by the language head or preserved as a vector for downstream calls. This procedure yields an interpretable aggregation that privileges modality/task match and component confidence, while remaining robust when evidence is distributed across modalities.

The graph composition model represents the execution for a complex query $Q$ as a set of nodes, where each node corresponds to a modality-specific agent responsible for handling a particular subtask:

\begin{equation}
G(Q) = {N_i}_{i=1}^k, \quad \text{where } N_i = (m_i, c_i, d_i)
\end{equation}

\begin{equation}
Y = \text{MoE}\left(\bigcup_{i=1}^{k} \text{Out}(N_i)\right)
\end{equation}
Here, $\text{MoE}(\cdot)$ denotes a weighted fusion of component outputs, where non-negative mixture weights sum to one and are derived from modality/task alignment and component confidence signals (cf.\ Section~\ref{sec:routing-orchestration}).

Each node $N_i$ is defined by the following components:
\begin{itemize}
\item $m_i$ — the {modality handler}, indicating the type of input processed (e.g., \textit{Image}, \textit{Text}, \textit{Document}).
\item $c_i$ — the {capability invoked}, specifying the operation to be performed (e.g., \textit{summarization}, \textit{object detection}, \textit{table parsing}).
\item $d_i$ — the set of {downstream dependencies}, capturing which other nodes consume this node’s output.
\end{itemize}
The final response $Y$ is computed by aggregating outputs from all nodes using a Mixture-of-Experts mechanism.

\subsubsection{Context Agent and Memory Management}
The \emph{Context Agent} governs a layered memory system that ensures coherent multi-turn interactions, preserves modality grounding, and enables context-aware model routing. Memory handling is dynamically adapted to the query path—whether through a single LLM, a decomposed agent sequence, or a multimodal fusion via Mixture-of-Experts. Table~\ref{tab:memory_layers} lists the memory layers and their roles.

\begin{table}[h]
  \centering
  \caption{Memory layers in the Context Agent and their roles.}
  \label{tab:memory_layers}
  \begin{tabular}{p{0.26\textwidth} p{0.67\textwidth}}
    \toprule
    \textbf{Memory Layer} & \textbf{Description} \\
    \midrule
    Short-Term Memory & Maintains a limited window of recent exchanges across modalities to support reactive follow-ups and continuity. \\
    Full Conversation History & A persistent, searchable log of the session, indexed semantically to enable fallback linking and long-horizon coherence. \\
    Module-Specific Memories & Isolated memory banks for each modality (e.g., audio, document, image) to ensure precise retrieval within the relevant interpretive domain. \\
    Relevant Query Context & Dynamically retrieves the most semantically similar past interactions using embedding-based similarity scoring. \\
    Compressed Context & A distilled summary representation capturing the global intent and entity trajectory of the session—used in context-constrained execution paths. \\
    \bottomrule
  \end{tabular}
\end{table}

These memory sources are combined selectively based on relevance to the current query. For instance, a follow-up text referencing a prior image will access both compressed history and the image memory bank—without pulling unrelated traces. This minimizes hallucinations and ensures efficient token usage.

Relevance of each candidate memory layer \(L_i\) is scored using:
\begin{equation}
\setlength{\abovedisplayskip}{6pt}  
\setlength{\belowdisplayskip}{6pt}  
\mathrm{ContextScore}(L_i)
  = \theta_s\,R_i \;+\;\theta_t\,T_i \;+\;\theta_m\,M_i
\end{equation}
Here, \(R_i\) denotes the semantic relevance to the current query, \(T_i\) represents the recency of last access, and \(M_i\) is the modality alignment score. The parameters \(\theta_s\), \(\theta_t\), and \(\theta_m\) are tunable weights balancing these factors.

Weight calibration is guided by ablation studies on multi‑turn dialogue corpora. Semantic relevance is given dominant emphasis because embedding similarity correlates most strongly with response quality, a finding validated across thousands of conversational triples. Temporal recency carries moderate weight to preserve natural discourse flow, with decay functions tuned for coherence over extended turns. Modality alignment supplies fine‑grained adjustment that maintains cross‑modal context integrity. During long sessions, the weighting scheme adapts: as the effective context window expands, the relative weight of semantic relevance increases to counter topic drift, while recency weight decays proportionally.
Only layers exceeding a threshold ContextScore are merged into the prompt context. In multi-agent flows, each agent receives only the memory slice relevant to its task, enabling efficient and scoped parallel reasoning.

\subsubsection{Routing Feedback and Policy Adjustment}
We incorporate a continuous feedback mechanism to refine routing decisions and enhance long-term system performance. After each query execution, users may rate the output or flag issues such as inappropriate model selection, missing context, or subpar results.

Collected feedback informs an adaptive routing loop that handles two scenarios in practice. When an output is marked unsatisfactory, the system triggers an automatic fallback to a more capable model (e.g., GPT-4), re-executes the query, and logs both outcomes for comparative evaluation. When a routing error is reported—such as incorrect model or modality selection—the system records the originally selected model, modality tag, and subtask, reroutes the query through alternative model paths, and captures the resulting quality differentials.

This feedback-driven tuning mechanism allows us to adjust routing strategies over time—improving alignment with task requirements and reducing the need for manual system updates. While the current system uses rule-based feedback assimilation, future iterations may incorporate learned policy adjustments via reinforcement or reward modeling.

\subsection{Couplet Framework Integration}
\label{sec:couplet}
The {Couplet Framework} in our architecture enables the orchestration of traditional machine learning models (e.g., CLIP, YOLO, Tesseract~\cite{smith2007overview}) using lightweight Small Language Models (SLMs) for interpretation and task structuring. This design offloads high-computation modules—like feature detection and visual object classification—away from general-purpose LLMs, ensuring efficient query execution while preserving modular intelligence.

\subsubsection{Case Study: ChatGPT’s Video Analysis}
To concretely illustrate the typical multimodal workaround in current LLM systems, consider a user query such as: "\emph{Analyze this video and summarize its key content}." A standard centralized approach, such as ChatGPT, addresses this query by generating Python code to extract frames, process them, and summarize—rather than using any vision-native model:
\begin{center}
\noindent\fbox{%
    \parbox{0.95\linewidth}{%
\small
\texttt{import cv2}\\
\texttt{import matplotlib.pyplot as plt}\\
\texttt{cap = cv2.VideoCapture('clip.mp4')}\\
\texttt{fps = cap.get(cv2.CAP\_PROP\_FPS)}\\
\texttt{frame\_count = cap.get(cv2.CAP\_PROP\_FRAME\_COUNT)}\\
\texttt{duration = frame\_count / fps if fps else 0}\\
\texttt{width = cap.get(cv2.CAP\_PROP\_FRAME\_WIDTH)}\\
\texttt{height = cap.get(cv2.CAP\_PROP\_FRAME\_HEIGHT)}\\
\texttt{mid\_frame = int(frame\_count // 2)}\\
\texttt{cap.set(cv2.CAP\_PROP\_POS\_FRAMES, mid\_frame)}\\
\texttt{ret, frame = cap.read()}\\
\texttt{if ret:}\\
\texttt{\quad frame\_rgb = cv2.cvtColor(frame, cv2.COLOR\_BGR2RGB)}\\
\texttt{\quad plt.imshow(frame\_rgb)}\\
\texttt{\quad plt.axis('off')}\\
\texttt{\quad plt.title("Representative Frame")}\\
\texttt{\quad plt.show()}\\
\texttt{cap.release()}
}}
\end{center}
\noindent This “code-first” workaround works generically but bypasses mature, domain-specific models—such as YOLO for object detection,  CLIP \cite{radford2021learning} for semantic embedding, or OCR engines for document parsing—that represent the state of the art for their respective tasks. Historically, each perception domain relied on its own optimized models; but as general-purpose LLMs have become widespread across applications, these specialist tools have been increasingly under-utilized.

The {Couplet Framework} in our architecture reverses this trend. It elevates proven traditional models into first-class routing targets. The orchestration layer interprets user intent, dispatches structured calls, and composes final responses—while the domain-specific models execute the perceptual subtasks with minimal overhead.
{Importantly}, Couplet is not just a vision-specific hack. Its architecture generalizes to other perceptual tasks—such as OCR for structured documents, tabular form extraction in compliance workflows, and acoustic signal interpretation in audio analytics. 

\subsubsection{Operational Flow}
\begin{figure}[htbp]
\centering
\includegraphics[width=0.98\textwidth]{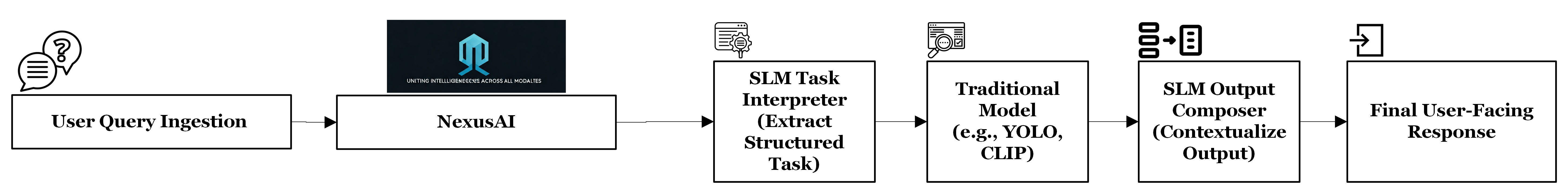}
\caption{\small\textit{Couplet Framework Execution Flow: Symmetric routing from user query to traditional models with SLM coordination.}}
\label{fig:couplet_flowchart}
\end{figure}

Lightweight coordination is achieved using intent parsers and schema mappers, which may involve SLMs when necessary. Figure~\ref{fig:couplet_flowchart} illustrates the Couplet pipeline. A user query is parsed by an SLM, structured into a task, delegated to a domain-specific model (e.g., YOLO, Tesseract), and reformatted into a final, human-readable response.

This architecture exemplifies our core philosophy: leverage existing domain-optimized tools where available, and allow lightweight language models to serve as intelligent, low-cost coordinators. Rather than relying on monolithic LLMs to solve perception tasks indirectly (e.g., through code generation), Couplet delegates these subtasks to well-established models—such as YOLO for detection or CLIP for semantics—and uses SLMs to interpret user queries and compose coherent outputs.

Operationally, Couplet follows a three-stage flow—SLM-based decomposition, traditional model execution, and SLM contextualization—applied sequentially. The SLM extracts a structured task representation from the natural language query (e.g., "Detect objects and label their function"); subsequently, the structured task is dispatched to a specialized model (e.g., YOLO for object detection, CLIP for semantic embedding, Tesseract for OCR); ultimately, the raw outputs return to the SLM, which reformulates them into a user-facing, semantically rich, and context-aware response.
This design preserves the performance advantages of domain-specific models and uses SLMs for lightweight coordination, showing that composition—not monolithic generation—can underpin scalable, cost-efficient pipelines.

\section{Evaluation and Results}
\label{sec:results}
We evaluated the performance of this architecture using large-scale samples from diverse standard benchmarks covering multiple modalities and task types: MMLU~\cite{hendrycks2020mmlu} for text reasoning, GSM8K~\cite{gsm8k_cobbe2021} for mathematical problem solving, MBPP~\cite{austin2021mbpp} for coding challenges, XSum~\cite{narayan2018don} and CNN/DailyMail~\cite{hermann2015teaching} for text summarization, VQA-v2~\cite{antol2015vqa} for visual question answering, FUNSD~\cite{jaume2019funsd} for document understanding, as well as audio transcription and video analysis prompts. We compare this architecture against a baseline {Always-Premium} strategy, where every query is handled by a single top-tier proprietary LLM (e.g., GPT-4). In the following, we present comprehensive results on routing accuracy, end-to-end answer quality across all these modalities and benchmarks, cost efficiency, system latency, and detailed analysis of failure modes and ablation studies.

\subsection{Routing and Classification Performance}
\label{sec:routing}
To stress-test the router, we compiled a diverse evaluation set of queries covering a wide range of domains and modalities. These included text-only tasks such as summarization (from XSum~\cite{narayan2018don}, CNN/DailyMail~\cite{hermann2015teaching}) and open-ended Q\&A, mathematical reasoning problems (GSM8K~\cite{gsm8k_cobbe2021}), coding challenges (MBPP~\cite{austin2021mbpp}), as well as queries involving images (VQA-v2~\cite{antol2015vqa} for visual QA, FUNSD~\cite{jaume2019funsd} for document understanding), and other modalities (audio transcription, video analysis prompts, image generation requests, etc.). We defined 13 fine-grained task categories to label this dataset, which covered both the query’s content and modality requirements. The categories were as follows: \texttt{math}, \texttt{coding}, \texttt{summarization\_writing}, \texttt{vision}, \texttt{document}, \texttt{image-gen}, \texttt{audio}, \texttt{video}, \texttt{text-moe} (follow-up text queries to non-text tasks), \texttt{object\_detect}, \texttt{complex}, \texttt{ambiguous}, and \texttt{general}.

To evaluate fine-grained routing, we constructed a balanced 13-way dataset via stratified sampling from established benchmarks—GSM8K~\cite{gsm8k_cobbe2021}, MBPP~\cite{austin2021mbpp}, XSum~\cite{narayan2018don}, CNN/DailyMail~\cite{hermann2015teaching}, VQA-v2~\cite{antol2015vqa}, and FUNSD~\cite{jaume2019funsd}—augmented with controlled generation for audio, video, and image-generation prompts. Category design covers simple–complex instances; \emph{complex} includes multi-step, cross-modal tasks requiring decomposition, \emph{ambiguous} probes decision boundaries, and \emph{text-moe} denotes textual follow-ups to prior non-text inputs. Labels were produced by two independent annotators under written guidelines, with disagreements resolved by consensus and GPT-4 arbitration; we measured inter-annotator agreement (Cohen’s $\kappa$) and re-annotated items with low agreement. This protocol yields category balance and realistic query diversity suitable for evaluating fine-grained routing performance.

The routing classifier was evaluated on its ability to correctly predict these categories and identify attached file(s) content, focusing on metrics such as overall routing accuracy, per-class precision/recall, classifier $F_{1}$ score, and attachment detection performance. Table~\ref{tab:routing} summarizes the key routing metrics. On the overall routing task (i.e., correctly selecting the appropriate model or modality for each query), we achieved an accuracy of {92.3\% ± 1.1\%}, with a macro-precision of 0.91 and macro-recall of 0.93. For the challenging fine-grained classification test with 13 task-specific categories (where each query’s true category was carefully annotated via human review and GPT-4 arbitration), the router still attained {86.78\%} accuracy, closely matching the performance of an expert (GPT-4) classifier on the same queries. The classifier’s overall $F_{1}$ score was 0.89.

Importantly, the system’s attachment detector—which is responsible for recognizing when an image, document, audio, or other file is provided with the query. In our evaluation set, the attachment detector achieved high performance on our evaluation dataset (precision and recall both 1.00 on the test set). While this result is encouraging, it should be interpreted as dataset-specific rather than a guaranteed outcome on all future inputs.

The only source of routing error or ambiguity arises in the case of text queries that serve as follow-ups to previous non-text tasks (the \texttt{text-moe} category). Here, the router must infer from context whether the current text is referencing an earlier non-text input, which can occasionally result in misclassification. For these follow-up queries, the detection accuracy is approximately {90\%}, with most failures arising from subtle or ambiguous textual references to prior modalities.
These results show that the router reliably recognizes the query's required modality and complexity, providing a strong foundation for subsequent model selection.

\begin{table}[h]
  \centering
  \caption{Routing and classification performance metrics.}
  \label{tab:routing}
  \sisetup{table-format=2.2}
  \begin{tabular}{l S[table-format=2.2] S[table-format=1.2] S[table-format=1.2]}
    \toprule
    \textbf{Metric} & \textbf{Value (\%)} & \textbf{Precision} & \textbf{Recall} \\
    \midrule
    Routing Accuracy (coarse)    & 92.3  & 0.91 & 0.93 \\
    Fine-Grained Accuracy (13-way) & 86.78 & \multicolumn{1}{c}{--} & \multicolumn{1}{c}{--} \\
    Classifier $F_{1}$–score     & 89.0  & \multicolumn{1}{c}{--} & \multicolumn{1}{c}{--} \\
    Attachment Detection Rate    & 100.0 & 1.00 & 1.00 \\
    Follow-up Detection Accuracy & 90.0  & 0.90 & 0.90 \\
    \bottomrule
  \end{tabular}
\end{table}

\subsection{End-to-End Accuracy and Cost Efficiency}
\label{sec:accuracy_cost}
We next evaluate end-to-end answer quality and cost relative to the Always-Premium baseline (AP), which routes every query to a single large multimodal model. Unless otherwise noted, reported metrics are mean values; where applicable, we also provide variance to indicate stability across subsets.

Table~\ref{tab:output_cost} reports the accuracy achieved on representative benchmarks, along with the normalized compute cost as a percentage of the Always-Premium scenario. On the {MMLU} \cite{hendrycks2020mmlu} language understanding benchmark—covering a broad range of knowledge domains in a QA format—we attained {88.5\%} accuracy, outperforming the Always-Premium baseline (GPT-4 serving all queries) which scored 84.2\%. Similarly, on the {VQA-v2} \cite{antol2015vqa} visual question-answering task, we achieved a {93.2\%} accuracy compared to 89.7\% for the baseline.

In addition to benchmark accuracy, we measured how closely our outputs match those from the Always-Premium model on open-ended prompts. Using the Always-Premium system’s responses as reference outputs, we computed lexical and semantic similarity metrics for a set of random user prompts. The average TF-IDF cosine similarity between our response and the reference was 0.63 (indicating moderately high word-level overlap), while the average BERT-based cosine similarity was 0.93 – indicating a very high semantic equivalence. In fact, {94\%} of our responses had a BERT similarity above 0.8, meaning the answers retained very close meaning to the reference outputs. This confirms that these routing decisions do not significantly degrade answer quality. Meanwhile, the routing efficiency on these prompts was excellent: we handled about {96\%} of the queries using smaller cost-efficient models, needing to invoke a top-tier model (GPT-4) for only the remaining 4\%. In other words, even while maintaining high answer fidelity, the system was able to resolve the vast majority of questions with cheaper models.

\begin{table}[h]
  \centering
  \caption{Our End-to-end benchmark accuracy vs.\ an Always-Premium baseline, and relative compute cost.}
  \label{tab:output_cost}
  \sisetup{table-format=3.1}
  \begin{tabular}{l S[table-format=2.1] S[table-format=2.1] S[table-format=3.0]}
    \toprule
    \textbf{Benchmark}                             & \textbf{Baseline Accuracy (\%)} & \textbf{Our Accuracy (\%)} & \textbf{Cost (\% of baseline)} \\
    \midrule
    MMLU (Language reasoning)             & 84.2                     & 88.5 ± 1.8\%                     &  70                       \\
    VQA-v2 (Visual QA)                    & 89.7                     & 93.2 ± 2.3\%                     &  68                       \\
    Fine-Grained Routing Eval (13-way)    & \multicolumn{1}{c}{--}  & 86.8                     &  42                       \\
    \bottomrule
  \end{tabular}
\end{table}

Simultaneously, we offer substantially improved cost-efficiency. The inference cost was reduced to roughly one-third of the Always-Premium approach. In a focused fine-grained routing evaluation, the cost savings were even more pronounced: we incurred only about 42\% of the baseline cost (a 58\% reduction), reflecting the precise routing and effective load distribution among models.

\begin{figure}[htbp]
  \centering
  \begin{tikzpicture}
    \begin{axis}[
      ybar stacked,
      bar width=28pt,
      width=0.65\textwidth,
      height=0.37\textwidth,
      ylabel={Share of Total Cost (\%)},
      symbolic x coords={All GPT-4, Our Routing},
      xtick=data,
      enlarge x limits=0.3,
      ymin=0, ymax=100,
      nodes near coords,
      every node near coord/.append style={font=\scriptsize},
      legend style={at={(1.04,1)}, anchor=north west, font=\small, cells={anchor=west}},
      title={Relative Cost Breakdown for Prompt Routing}
    ]
      \addplot+[fill=red!60] coordinates {(All GPT-4,100) (Our Routing,4)};
      \addplot+[fill=green!60] coordinates {(All GPT-4,0) (Our Routing,55)};
      \addplot+[fill=blue!60] coordinates {(All GPT-4,0) (Our Routing,30)};
      \addplot+[fill=orange!60] coordinates {(All GPT-4,0) (Our Routing,11)};
      \legend{GPT-4, Qwen 2.5, Llama 3.1, Mixtral}
    \end{axis}
  \end{tikzpicture}
  \caption{\textit{Relative share of total processing cost for different models under two scenarios: (1) all queries routed to GPT-4 ({All GPT-4}), and (2) our routing approach.}}
  \label{fig:cost_breakdown}
\end{figure}
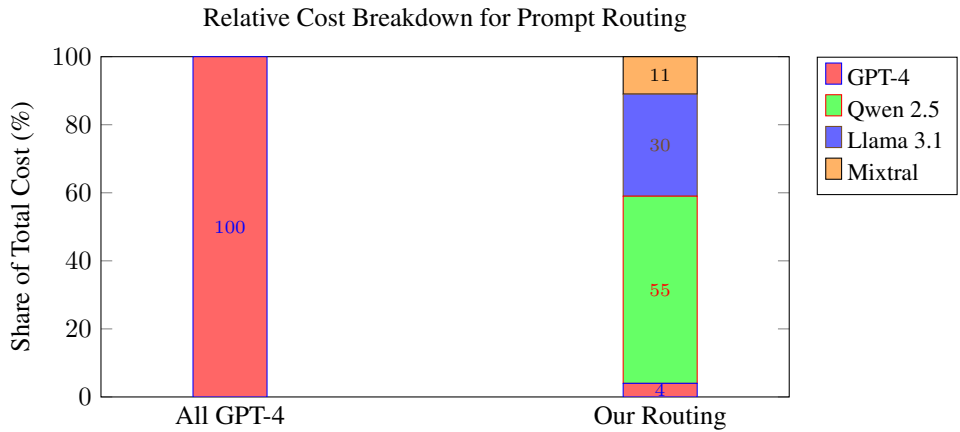

The substantial cost savings stem from the fact that a majority of queries do not require an expensive model. In our experiments, approximately {72\%} of all queries were fully handled by efficient open-source models (using models such as Mistral~\cite{mistral7b_2023} or LLaVA for vision), because these queries were relatively straightforward or fell into domains where smaller models performed adequately. Only about 28\% of queries needed to be escalated to a closed-source model (such as GPT-4 or image generation tools\footnote{Some references to image generation capabilities were created with the assistance of DALL-E 3 (OpenAI, 2023).}) – typically those that involved complex reasoning, ambiguous instructions, or specialized capabilities like high-fidelity image generation. Table~\ref{tab:routing-distribution} shows the distribution of queries between the open-source paths and the premium path. By routing the bulk of requests to local models, we were able to cut the overall usage of paid API calls by more than half, while still maintaining or even improving answer accuracy as noted above. A breakdown of cost by model type further highlights these savings: Figure~\ref{fig:cost_breakdown} compares the total cost of routed prompts for each model category, showing that we slash cost across the board by utilizing efficient models like Qwen-2.5, LLaMA 3.1, Mistral, etc., for most queries and only rarely relying on the expensive GPT-4.

As shown in Figure~\ref{fig:cost_breakdown}, the overwhelming majority of processing cost in the Always-Premium deployment is attributed solely to GPT-4. In contrast, with our approach’s routing, the cost is spread across several more efficient models: Qwen 2.5 accounts for over half of the total cost, while Llama 3.1 and Mixtral together handle more than 40\%. Only about 4\% of the total cost is due to GPT-4, as its use is now reserved for the most complex queries. This distribution illustrates how we achieve substantial cost savings by utilizing fast, cost-effective models for most queries, and only rarely escalating to premium models, while still meeting or exceeding baseline accuracy. The figure highlights the strategic load balancing and efficiency achieved by model-aware routing.
\begin{table}[h]
  \centering
  \caption{Query and cost share for open-source and premium models.}
  \label{tab:routing-distribution}
  \begin{tabular}{l c c}
    \toprule
    \textbf{Routing Path} & \textbf{Query Share (\%)} & \textbf{Cost Share (\%)} \\
    \midrule
    Open-Source (Ollama)       & 72 & 30 \\
    Premium (GPT-4\textsubscript{o}, DALL·E~3) & 28 & 70 \\
    \bottomrule
  \end{tabular}
\end{table}

\begin{figure}[htbp]
  \centering
  \begin{tikzpicture}
    \begin{axis}[
      width=0.8\textwidth,
      height=0.36\textwidth,
      xlabel={Relative Cost (\% of Always-Premium)},
      ylabel={Accuracy (\%)},
      xmin=60, xmax=102, ymin=80, ymax=100,
      grid=both,
      enlarge x limits={abs=2},
      enlarge y limits={abs=1},
      legend style={at={(0.5,-0.25)}, anchor=north, legend columns=2, font=\small},
      font=\small
    ]
      \addplot[mark=*, blue] coordinates {(100,84.2) (100,89.7)};
      \addplot[mark=square*, red] coordinates {(70,88.5) (68,93.2)};
      \legend{Always-Premium, Our Approach}

      \node[font=\small\bfseries\color{blue}, anchor=east, xshift=-2pt] at (axis cs:100,84.2) {AP MMLU};
      \node[font=\small\bfseries\color{blue}, anchor=east, xshift=-2pt] at (axis cs:100,89.7) {AP VQA};
      \node[font=\small\bfseries\color{red}, anchor=west, xshift=2pt] at (axis cs:70,88.5) {OA MMLU};
      \node[font=\small\bfseries\color{red}, anchor=west, xshift=2pt] at (axis cs:68,93.2) {OA VQA};
    \end{axis}
  \end{tikzpicture}
  \caption{\textit{
    Cost vs. accuracy trade-off on text and vision benchmarks. Our Approach (OA) outperforms Always-Premium (AP) in both dimensions. All costs are shown relative to the total compute cost in the Always-Premium baseline (normalized to 100\%). No specific currency is used.}
  }
  \label{fig:cost_accuracy}
\end{figure}
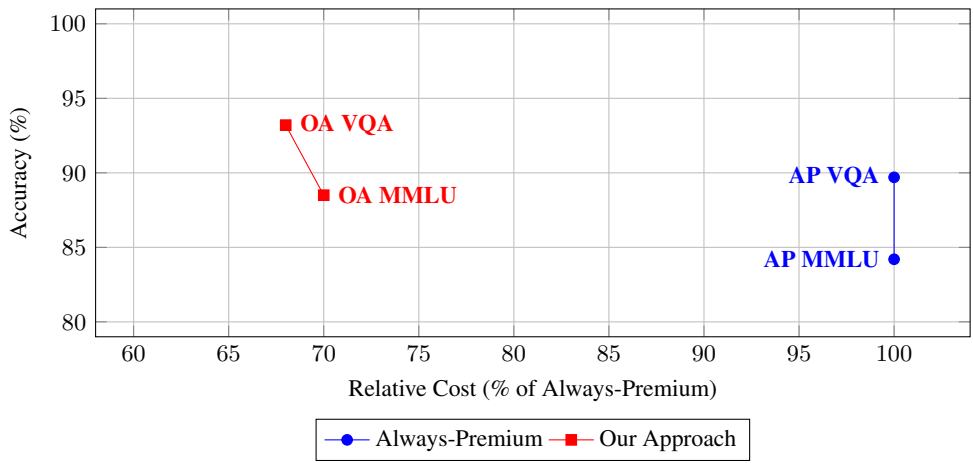

The trade-off between accuracy and cost is illustrated in Figure~\ref{fig:cost_accuracy}. Each point on the plot compares the Always-Premium deployment (at 100\% relative cost) to our deployment (at reduced cost) for a given benchmark. For both the {MMLU} and {VQA-v2} tasks, we achieved higher accuracy than the baseline while using significantly less computational cost. For example, on VQA, we reached about 93\% accuracy at only 68\% of the cost, whereas the baseline (GPT-4 alone) was 90\% accurate at full cost. This highlights that intelligent routing can push the efficiency frontier, obtaining better results with fewer resources by leveraging specialized models where appropriate.

We evaluated this architecture in terms of runtime efficiency, measuring the end-to-end latency per query and the total throughput (queries processed per second). Because we introduce a routing stage before generating an answer, there is some added overhead, but the benefits of using faster models for many queries can offset this.

Table~\ref{tab:latency} reports the average response times for the Always-Premium system versus this architecture, broken down by modality, as well as the median routing-only latency. For purely textual queries, the average end-to-end response time under the Always-Premium setup (GPT-4 answering everything) was about 512~ms, whereas our approach responded in 419~ms on average—a reduction of roughly 18\% in latency. A similar improvement was observed for vision (image-based) queries: 645~ms with the baseline vs.\ 530~ms with our approach (about 18\% faster). These speedups are achieved because we often utilize smaller, faster models (with far fewer parameters) when possible. For example, a local LLaMA-2 model can answer a straightforward text question faster than making an API call to a large model, and a lightweight vision model can handle a simple image query faster than a generalist multimodal model. The routing decision itself is very fast: the cascade of keyword detection, syntax parsing, Valpy scoring, and Semantic Embeddings arbitration takes under 300~ms on average, which is negligible compared to typical LLM generation times for non-trivial queries. Moreover, many components of the routing (like the keyword and syntax checks) run in parallel or have minimal cost.

In terms of throughput, we were able to handle about 20\% more queries per second than the baseline system. In a setting where multiple requests are processed concurrently, the Always-Premium approach maxed out at roughly 45 queries per second (qps) given our hardware and the rate-limited API, while our approach achieved about 54 qps. This improvement is mainly due to the workload being distributed: many queries are answered by local models (which can run in parallel on our own infrastructure without strict rate limits), reducing the bottleneck on the single premium model. It demonstrates that the routing approach not only cuts cost but also improves the scalability and responsiveness of the system.
\begin{table}[h]
  \centering
  \caption{Latency and throughput comparison. Response times are averaged over the evaluation queries for each modality.}
  \label{tab:latency}
  \sisetup{table-format=3.1}
  \begin{tabular}{l S[table-format=3.1] S[table-format=3.1] S[table-format=2.1]}
    \toprule
    \textbf{Metric}                  & \textbf{Always-Premium (ms)} & \textbf{Our Approach (ms)} & \textbf{Improvement (\%)} \\
    \midrule
    Avg.\ response time (text)       & 512 & 419 & 18.2 \\
    Avg.\ response time (vision - couplet)     & 645 & 530 & 17.9 \\
    Routing latency only             & {--}  & 290 & {--}   \\
    Throughput (queries per second)  & 45  & 54  & 20.0 \\
    \bottomrule
  \end{tabular}
\end{table}

\section{Conclusion}
The proliferation of large language models has created a fundamental tension in AI deployment: while models like GPT-4, Claude, and Gemini demonstrate exceptional capabilities across diverse tasks, their computational costs render uniform deployment economically prohibitive for large-scale applications. This challenge is particularly acute in multimodal scenarios where queries span text, vision, audio, video, and document modalities, each requiring specialized processing capabilities that general-purpose models handle inefficiently.

In this work, we have presented a modular, cost-aware routing framework that addresses this fundamental challenge through intelligent orchestration of specialized models across multiple modalities. Our comprehensive evaluation demonstrates that this approach achieves superior performance compared to monolithic deployment strategies while delivering substantial cost reductions. The significance of these findings extends beyond mere performance metrics. Our architecture's dynamic routing, which correctly identifies optimal processing paths for over 92\% of heterogeneous queries, demonstrates that intelligent orchestration can fundamentally reshape how we deploy AI systems at scale.

While our evaluation demonstrates the effectiveness of our architecture across multiple benchmarks and modalities, several limitations warrant acknowledgment. The current framework relies on a fixed 13-category classification scheme for task routing, which may not capture the full spectrum of emerging query types as AI applications evolve. Additionally, the complexity of orchestrating multiple specialized models introduces potential points of failure that require robust monitoring and fallback mechanisms. The framework's performance is also dependent on the quality and availability of specialized models for each domain, which may vary across different application contexts.

Future research directions include extending the routing agent with continual learning capabilities to enable online adaptation to emerging query patterns, integrating additional modalities such as real-time sensor streams and interactive code execution environments, and exploring tighter user-in-the-loop feedback mechanisms for dynamic adjustment of routing thresholds.

This work represents a principled approach to addressing the cost-performance trade-offs inherent in modern AI deployment, offering a scalable blueprint for building multimodal AI services that meet enterprise-grade requirements for accuracy, efficiency, and cost control. By demonstrating that intelligent orchestration of specialized components can outperform monolithic approaches while delivering substantial economic benefits, this work contributes to a more sustainable and accessible future for AI system deployment across diverse application domains.
\printbibliography
\end{document}